\documentclass[10pt,twocolumn,letterpaper]{article}

\usepackage{cvpr}
\usepackage{times}
\usepackage{graphicx}
\usepackage{amsmath}
\usepackage{amssymb}
\usepackage[american]{babel}
\usepackage{multirow}
\usepackage{subcaption}
\usepackage{algorithm}
\usepackage{algorithmicx}
\graphicspath{{figure/}}

\usepackage[pagebackref=true,breaklinks=true,colorlinks,bookmarks=false]{hyperref}

\cvprfinalcopy 

\def\cvprPaperID{6555} 

\ifcvprfinal\fi

\begin{document}

\title{Rethinking the Route Towards Weakly Supervised Object Localization}

\author{Chen-Lin Zhang \qquad Yun-Hao Cao \qquad Jianxin Wu\thanks{This research was partially supported by the National Natural Science Foundation of China (61772256, 61921006). J. Wu is the corresponding author.}\\
National Key Laboratory for Novel Software Technology \\
Nanjing University, Nanjing, China\\
{\tt\small \{zhangcl,caoyh\}@lamda.nju.edu.cn, wujx2001@nju.edu.cn}
}

\maketitle

\begin{abstract}
   Weakly supervised object localization~(WSOL) aims to localize objects with only image-level labels. Previous methods often try to utilize feature maps and classification weights to localize objects using image level annotations indirectly. In this paper, we demonstrate that weakly supervised object localization should be divided into two parts: the class-agnostic object localization and the object classification. For class-agnostic object localization, we should use class-agnostic methods to generate noisy pseudo annotations and then perform bounding box regression on them without class labels. We propose the pseudo supervised object localization~(PSOL) method as a new way to solve WSOL. Our PSOL models have good transferability across different datasets without fine-tuning. With the generated pseudo bounding boxes, we achieve 58.00\% localization accuracy on ImageNet and 74.97\% localization accuracy on CUB-200, which have a large edge over previous models.
\end{abstract}

\section{Introduction}
Deep convolutional neural networks have achieved enormous success in various computer vision tasks, such as classification, localization and detection. However, current deep learning models need a large number of accurate annotations, including image-level labels, location-level labels~(bounding boxes and key points) and pixel-level labels~(per pixel class labels for semantic segmentation). Many large-scale datasets are proposed to solve this problem~\cite{imagenetijcv2015,cocoeccv2014,cityscapescvpr2016}. However, models pre-trained on these large-scale datasets cannot be directly applied to a different task due to the differences between source and target domains. 

To relax these restrictions, weakly supervised methods are proposed. Weakly supervised methods try to perform detection, localization and segmentation tasks with only image-level labels, which are relatively easy and cheap to obtain. Among these tasks, weakly supervised object localization~(WSOL) is the most practical task since it only needs to locate the object \textit{with a given class label}. Most of these WSOL methods try to enhance the localization ability of classification models to conduct WSOL tasks~\cite{hideandseekiccv2017,acolcvpr2018,spgeccv2018,adlcvpr2019,cutmixiccv2019} using the class activation map~(CAM)~\cite{camcvpr2016}. 

\begin{figure*}[t]
	\begin{center}
		\includegraphics[width=1.6\columnwidth]{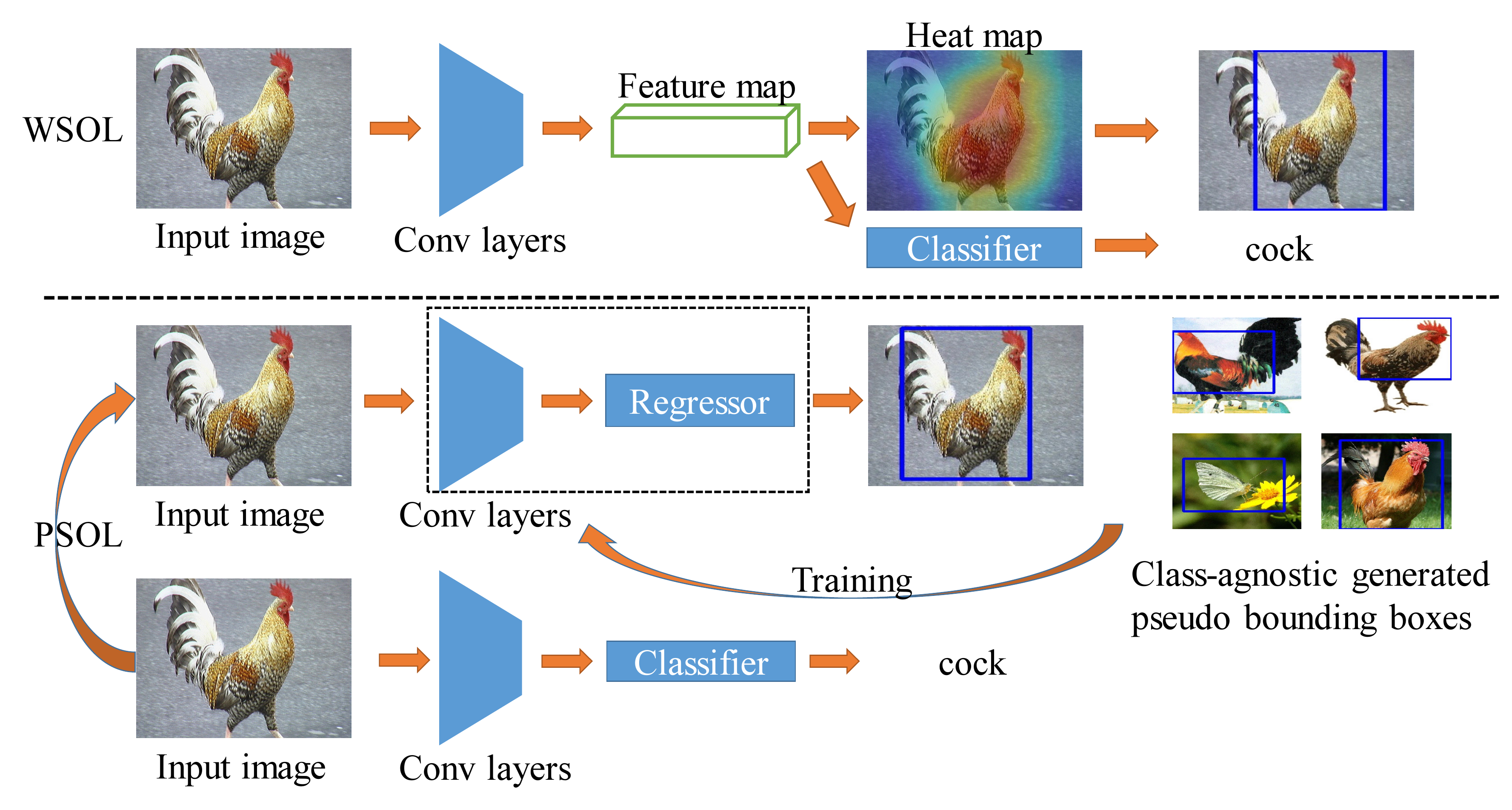}
	\end{center}
	\caption{Overall pipeline of previous WSOL methods~(top) and our proposed PSOL~(bottom). Previous WSOL methods need the final feature map to generate bounding boxes \emph{implicitly}. However, PSOL first generates inaccurate bounding boxes using class-agnostic methods, then perform bounding box regression to predict the bounding box \emph{explicitly}.}
	\label{fig:overall}
\end{figure*}

However, in this paper, through ablation studies and experiments, we demonstrate that the localization part of WSOL should be \textit{class-agnostic}, which is not related to classification labels. Based on these observations, we advocate a paradigm shift which divides WSOL into two independent sub-tasks: the class-agnostic object localization and the object classification. The overall pipeline of our method is in Fig.~\ref{fig:overall}. We name this  novel pipeline as \textit{Pseudo Supervised Object Localization~(PSOL)}. We first generate pseudo groundtruth bounding boxes based on the class-agnostic method deep descriptor transformation~(DDT)~\cite{ddtpr2019}. By performing bounding box regression on these generated bounding boxes, our method removes restrictions on most WSOL models, including the restrictions of allowing only one fully connected layer as classification weights~\cite{camcvpr2016} and the dilemma between classification and localization~\cite{hideandseekiccv2017,adlcvpr2019}.

We achieve state-of-the-art performances on ImageNet-1k~\cite{imagenetijcv2015} and CUB-200~\cite{cubtech2011} combining the results of these two independent sub-tasks, obtaining a large edge over previous WSOL models~(especially on CUB-200). With classification results of the recent EfficientNet~\cite{efficientneticml2019} model, we achieve 58.00\% Top-1 localization accuracy on ImageNet-1k, which significantly outperforms previous methods.

We summarize our contributions as follows.
\begin{itemize}
	\item We show that weakly supervised object localization should be divided into two independent sub-tasks: the class-agnostic object localization and the object classification. We propose PSOL to solve the drawbacks and problems in previous WSOL methods.
	\item Though generated bounding boxes are noisy, we argue that we should directly optimize on them \emph{without using class labels}. With the proposed PSOL, we achieve 58.00\% Top-1 localization accuracy on ImageNet-1k and 74.97\% Top-1 localization accuracy on CUB-200, which is far beyond previous state-of-the-art.
	\item Our PSOL method has good localization transferability across different datasets \emph{without any fine-tuning}, which is significantly better than previous WSOL models. 
\end{itemize}

\section{Related Works}
Convolutional neural networks~(CNN), since the success of AlexNet~\cite{alexnetnips2012}, have been widely applied in many areas of computer vision, including object localization and object detection tasks. We will briefly review detection and localization with full supervision and weak supervision in this section.

\subsection{Fully Supervised Methods}
After the success of AlexNet~\cite{alexnetnips2012}, researchers tried to adopt CNN to conduct object localization and detection. The pioneering work OverFeat~\cite{overfeaticlr2014} tried to use sliding window and multi-scale techniques to conduct classification, localization and detection within a single network. VGG-Net~\cite{vggiclr2014} adds a per-class regression and model ensemble to enhance the prediction result of localization. 

Object detection is another task that can generate bounding boxes and labels simultaneously. R-CNN~\cite{rcnncvpr2014} and Fast-RCNN~\cite{fastrcnniccv2015} use the selective search~\cite{selectivesearchijcv2013} to generate candidate regions and then use CNN to classify them. Faster-RCNN~\cite{fasterrcnnnips2015} proposes a two-stage network: the region proposal network~(RPN) for generating regions of interest~(ROI), then the R-CNN module to classify them and localize the object in the region. These popular two-stage detectors are widely used in detection tasks. YOLO~\cite{yolov1cvpr2016} and SSD~\cite{ssdeccv2016} are one stage detectors with carefully designed network structures and anchors. Recently, some anchor-free detectors are proposed to mitigate the anchor problem in common detectors like CornerNet~\cite{cornetneteccv2018} and CenterNet~\cite{centernetarxiv2019}.

However, all these methods need massive, detailed and accurate annotations. Annotations in real-world tasks are expensive and sometimes even hard to get. So we need some other methods to perform object localization tasks without requiring many exact labels.
\subsection{Weakly Supervised Methods}
Weakly supervised object localization~(WSOL) learns to localize the object with only image-level labels. It is more attractive since image-level labels are much easier and cheaper to obtain than object level labels. Weakly supervised detection~(WSOD) tries to give the object location and category simultaneously when training images only have image-level labels. 

WSOL has the assumption that there is only one object of the specific category in the whole image. Based on this assumption, many methods are proposed to push the limit of WSOL.~\cite{camcvpr2016} first generates class activation maps with the global average pooling layer and the final fully connected layer~(weights of the classifier). Grad-CAM~\cite{gradcamiccv2017} uses gradients rather than output features to generate more accurate class response maps. Besides these methods which focus on improving class response maps, some other methods try to make the classification model more suitable for localization tasks. HaS~\cite{hideandseekiccv2017} tries to randomly erase some 
regions in the input image to force the network to be meticulous for WSOL. ACoL~\cite{acolcvpr2018} uses two parallel classifiers with dynamic erasing and adversarial learning to discover complementary object regions more effectively. SPG~\cite{spgeccv2018} generates Self-Produced Guidance masks to localize the whole object. ADL~\cite{adlcvpr2019} proposes the importance map and the drop mask, with a random selection mechanism to achieve a balance between classification and localization. 

WSOD does not have the \textit{one object in one class} restriction. However, WSOD often needs methods to generate region proposals like selective search~\cite{selectivesearchijcv2013} and edge boxes~\cite{edgeboxeccv2014}, which will cost much computation resources and time. Furthermore, current WSOD detectors use high resolution inputs to output the bounding boxes, leading to  heavy computational burdens. Thus, most WSOD methods are difficult to be applied to large-scale datasets.

\section{Methodology}

In this section, we will mainly discuss the drawbacks of the current WSOL pipeline and propose our \textit{pseudo supervised object localization~(PSOL)}.

\subsection{A paradigm shift from WSOL to PSOL}

Current WSOL nethods can generate the bounding box with a given class label. However, the community have identified serious drawbacks of this pipeline.
\begin{itemize}
	\item The learning objective is indirect, which will hurt the model's performance on localization tasks. HaS~\cite{hideandseekiccv2017} and ADL~\cite{adlcvpr2019} show that localization is not compatible with classification when only having a single CNN model. Localization tries to localize the whole object while classification tries to classify the object. The classification model often tries to localize only the most discriminative part of the object in an image.
	\item Offline CAM~\cite{camcvpr2016} has the thresholding parameter and needs to store the three-dimensional feature map for further computation. The thresholding value is tricky and hard to determine. 
\end{itemize}

Those drawbacks make WSOL hard to apply to real-world applications. 

Encouraged by the class-agnostic process that generats regions of interest~(ROI) in selective search~\cite{selectivesearchijcv2013} and Faster-RCNN~\cite{fasterrcnnnips2015}, we divide WSOL into two sub-tasks: the class-agnostic object localization and the object classification. Based on these two sub-tasks, we propose our PSOL method. PSOL directly optimizes the localization model on explicitly generated pseudo ground-truth bounding boxes. Hence, it removes the restrictions and drawbacks illustrated in previous WSOL methods, and it is a paradigm shift for WSOL.

\subsection{The PSOL Method}

The general framework of our PSOL is in Algorithm~\ref{alg:PSOL}. We will introduce our PSOL step by step. We will discuss the details of generating pseudo groundtruth bounding boxes in Sec~3.2.1, then the localization method used in our model in Sec~3.2.2. For the classification method, we use pre-trained models in the computer vision community directly.

\begin{algorithm}[t]
	
	\caption{Pseudo Supervised Object Localization}
	
	\begin{algorithmic} [1]
		\Statex {\textbf{Input}: Training images $I_{tr}$ with class label $L_{tr}$}
		\Statex {\textbf{Output}: Predicted bounding boxes $b_{te}$ and class labels $L_{te}$ on testing images $I_{te}$}
		\State \quad Generate pseudo bounding boxes $\tilde{b}_{tr}$ on $I_{tr}$ 
		\State \quad Train a localization CNN $F_{loc}$ on $I_{tr}$ with $\tilde{b}_{tr}$ 
		\State \quad Train a classification CNN $F_{cls}$ on $I_{tr}$ with $L_{tr}$
		\State \quad Use $F_{loc}$ to predict $b_{te}$ on $I_{te}$
		\State \quad Use $F_{cls}$ to predict $L_{te}$ on $I_{te}$
		\State \textbf{Return:} $b_{te}$, $L_{te}$
	\end{algorithmic}
	\label{alg:PSOL}
\end{algorithm}

\subsubsection{Bounding Box Generation} The critical difference between WSOL and our PSOL is the generation of pseudo bounding boxes for training images. Detection is a natural choice for this task since detection models can provide bounding boxes and classes directly. However, the largest dataset in detection only has 80 classes~\cite{cocoeccv2014}, and it cannot provide a general object localizer for datasets with many classes such as ImageNet-1k. Furthermore, current detectors like Faster-RCNN~\cite{fasterrcnnnips2015} need substantial computation resources and large input image sizes~(like shorter side=600 when testing). These issues prevent detection models from being applied to generate bounding boxes on large-scale datasets.

Without detection models, we can try some localization methods to output bounding boxes for training images directly. Some weakly and co-supervised methods can generate noisy bounding boxes, and we will give a brief introduction to them.

\textbf{WSOL methods.}
Existing WSOL methods often follow this pipeline to generate the bounding box for an image. First the image $I$ is feed into the network $F$, then the final feature map~(often the output of the last convolutional layer) $G$
is generated: $ G\in \mathbb{R}^{h\times w \times d} = F(I)$, where $h,w,d$ are the height, width and depth of the final feature map. Then, after global average pooling and the final fully connected layer, the label $L_{pred}$ is produced. According to the predicted label $L_{pred}$ or the ground truth label $L_{gt}$, we can get the class specific weights in the final fully connected layer $W \in \mathbb{R}^d$. Then each spatial location of $G$ is channel-wise weighed and summed to get the final heat map $H$ for the specific class: $H_{i,j} = \sum_{k=1}^{d}{G_{i,j,k} W_{k}}$. Finally, $H$ is upsampled to the original input size, and thresholding is applied to generate the final bounding box.

\textbf{DDT recap.}
Some co-supervised methods can also have good performances on localization tasks. DDT has good performance and little computational resource requirement among these co-supervised methods. So we use DDT~\cite{ddtpr2019} as an example. Here is a brief recap of DDT.
Given a set of images $S$ with $n$ images, where each image $I \in S$ has the same label, or contains the same object in the image. With a pre-trained model $F$, the final feature map is also generated: $ G\in \mathbb{R}^{h\times w \times d} = \mathbb{R}^{hw\times d} = F(I)$. Then these feature maps are gathered together into a large feature set: $G_{all} \in \mathbb{R}^{n \times hw \times d} = \mathbb{R}^{nhw\times d}$. Principal component analysis~(PCA)~\cite{pca1901} is applied along the depth dimension. After the PCA process, we can get the eigenvector $P$ with the largest eigenvalue. Then, each spatial location of $G$ is channel-wise weighed and summed to get the final heat map $H$: $H_{i,j} = \sum_{k=1}^{d}{G_{i,j,k} P_{k}}$. Then $H$ is upsampled to the original input size. Zero thresholding and max connected component analysis is applied to generate the final bounding box.
 
We will generate pseudo bounding boxes using both WSOL methods and the DDT method, and evaluate their suitability.

\subsubsection{Localization Methods}

After generating bounding boxes, we have (pseudo) bounding box annotations for each training image. Then it is natural to perform object localization with these generate boxes. As shown before, detection models are too heavy to handle this task. Thus, it is natural to perform bounding box regression. Previous fully supervised works~\cite{vggiclr2014,overfeaticlr2014} suggest two methods of bounding box regression: single-class regression~(SCR) and per-class regression~(PCR). PCR is strongly related to the class label. Since we advocate that localization is a class-agnostic rather than a class-aware task, we choose SCR for all our experiments. 

We follow previous work to perform bounding box regression~\cite{vggiclr2014}. Suppose the bounding box is in the $x,y,w,h$ format, where $x,y$ are the top-left coordinates of the bounding box and $w,h$ are the width and height of the bounding box, respectively. We first transfer $x,y,w,h$ into $x^*,y^*,w^*,h^*$ where $x^*=\frac{x}{w_{i}},y^*=\frac{y}{h_{i}}, w^* = \frac{w}{w_{i}}, h^* = \frac{h}{h_{i}} $, and $w_{i}$ and $h_{i}$ are the width and height of the input image, respectively. We use a sub-network with two fully connected layers and corresponding ReLU layers for regression. Finally, the outputs are sigmoid activated. We use the mean squared error loss~($\ell_2$ loss) for the regression task.

Step 2 and step 3 in Algorithm~\ref{alg:PSOL} may be combined, i.e., $F_{cls}$ and $F_{loc}$ can be integrated into a single model, which is jointly trained with classification labels and generated bounding boxes. However, we will show empirically that localization and classification models should be separated.

\section{Experiments}
\subsection{Experimental Setups}

\textbf{Datasets.} We evaluate our proposed method on two common WSOL datasets: ImageNet-1k~\cite{imagenetijcv2015} and CUB-200~\cite{cubtech2011}. The ImageNet-1k dataset is a large dataset with 1000 classes, containing 1,281,197 training images and 50,000 validation images. For training images, bounding box annotations are incomplete, and bounding box labels are complete for validation images. In this paper, \emph{we do not use any accurate training bounding box annotations.} In our experiments, we generate pseudo bounding boxes on training images by previous methods. The detailed ablation studies will be in Sec~5.1. We train all models on the generated bounding box annotations and classification labels and test them on the validation dataset. 

For the CUB-200 dataset, it contains 200 categories of birds with 5,994 training images and 5,794 testing images. Each image in the dataset has an accurate bounding box annotation. We follow the strategies on ImageNet-1k to train and test models.

\textbf{Metrics.} We use three metrics for evaluating our models: Top-1/Top-5 localization accuracy~(\textit{Top-1/Top-5 Loc}) and localization accuracy with known ground truth class~(\textit{GT-Known Loc}). They are following previous state-of-the-art methods~\cite{camcvpr2016,adlcvpr2019}: \textit{GT-Known Loc} is correct when given the ground truth class to the model, the intersection over union (IoU) between the ground truth bounding box and the predicted box is 50\% or more. \textit{Top-1 Loc} is correct when the Top-1 classification result and \textit{GT-Known Loc} are both correct. \textit{Top-5 Loc} is correct when given the Top-5 predictions of groundtruth labels and bounding boxes, there is one prediction which the classification result and localization result are both correct.

\textbf{Base Models.} We prepare several baseline models for evaluating our method on localization tasks: VGG16~\cite{vggiclr2014}, InceptionV3~\cite{inceptionv3cvpr2016}, ResNet50~\cite{resnetcvpr2016} and DenseNet161~\cite{densenetcvpr2017}. Previous methods try to enlarge the spatial resolution of the feature map~\cite{acolcvpr2018,spgeccv2018,adlcvpr2019}, we do not use this technology in our PSOL models. Previous WSOL methods need the classification weights to turn a 3D feature map into a 2D spatial heat map. However, in PSOL, we do not need the feature map for localization, our model will directly output the bounding box for object localization. For a fair comparison, we modified VGG16 into two versions: VGG-GAP and VGG16. VGG-GAP replaces all fully connected layers in VGG16 with GAP and a single fully connected layer, and VGG16 keeps the original structures in VGG16. For other models, we keep the original structure of each model. For regression, we use a two-layer fully connected network with corresponding ReLU layers to replace the last layer in original networks, as illustrated in Sec~3.2.2.

\textbf{Joint and Separate Optimization} In the previous section, we discussed the problem of joint optimization of classification and localization tasks. For ablating this issue, we prepare several models for each base model. For joint optimization models, we add a new bounding box regression branch to the model~(\texttt{-Joint} models), and then train this model with both generated bounding boxes and class labels simultaneously. For separate optimization models, we replace the classification part with the regression part~(\texttt{-Sep} models), then train these two models separately, i.e., localization models are trained with only generated bounding boxes while classification models are trained with only class labels. The hyperparameters are kept same for all models.

\subsection{Implementation Details}
We use the PyTorch framework with TitanX Pascal GPUs support. For all models, we use pre-trained classification weights on ImageNet-1k and fine-tune on target localization and classification tasks. 

For experiments on ImageNet-1k, the hyperparameters are set the same for all models: batch size 256, 0.0005 weight decay and 0.9 momentum. We will fine-tune all models with a start learning rate of 0.001. Added components~(like the regression sub-network) will have a larger learning rates due to the random initialization. We train 6 epochs on ImageNet and 30 epochs on CUB-200. For localization only tasks, we keep the learning rate fixed among all eppochs. The reason is that DDT generated bounding boxes are noisy, which contain many inaccurate or even totally wrong bounding boxes. The conclusion in~\cite{noisecvpr2018} shows that for noisy data, we should retain large learning rates. For classification related tasks~(including single classification and joint classification and localization tasks), we divide the learning rate by 10 every 2/10 epochs on ImageNet/CUB-200.

For testing models, we use ten crop augmentations on ImageNet to output results of the final classification following~\cite{acolcvpr2018} and~\cite{spgeccv2018} on ImageNet and single crop classification results on CUB200, and use single image inputs for all our localization results. We use the center crop techniques to get the image input, e.g., resize to 256$\times$256 then center crop to 224$\times$224 for most models except InceptionV3~(resize to 320x320 then center crop to 299$\times$299), following the setup in~\cite{adlcvpr2019, cutmixiccv2019}. For state-of-the-art classification models, we also follow the input size in their paper, e.g., 600 for EfficientNet-B7. 

Previous WSOL methods can provide multiple boxes for a single image with different labels. However, our SCR model can only provide one bounding box output for each image. Thus, we combine the output bounding box with Top-1/Top-5 classification outputs of baseline models~(\texttt{-Sep} models) or with outputs of the classification branch~(\texttt{-Joint} models) to get the final output to evaluate on test images.

For experiments on CUB-200, we change the batch size from 256 to 64, and keep other hyperparameters the same as ImageNet-1k.

\section{Results and Analyses}
In this section, we will provide empirical results, and perform detailed analyses on them.

\subsection{Ablation Studies on How to Generate Pseudo Bounding Boxes}

\begin{table}
	\caption{The \textit{GT-Known Loc} accuracy on the ImageNet-1k validation dataset of various weakly and co-supervised localization~(DDT) methods.}
	\label{table:gtknown results}
	\setlength{\tabcolsep}{2pt}
	\centering
	\begin{tabular}{|l|r|r|}
		\hline
		Model  & ImageNet-1k & CUB-200\\  \hline 
		VGG16-CAM~\cite{camcvpr2016} & 59.00 & 57.96\\
		VGG16-ACoL~\cite{acolcvpr2018} & 62.96 & 59.30\\
		SPG~\cite{spgeccv2018} & 64.69 & 60.50\\
		\hline
		DDT-ResNet50~\cite{ddtpr2019} & 59.92 & 72.39\\
		DDT-VGG16~\cite{ddtpr2019} & 61.41 & 84.55\\
		DDT-InceptionV3~\cite{ddtpr2019} & 51.87 & 51.80\\
		DDT-DenseNet161~\cite{ddtpr2019} & 61.92 & 78.09\\
		\hline
	\end{tabular}
\end{table}

Previous WSOL methods can generate bounding boxes with given ground truth labels. Some co-localization methods can also provide bounding boxes with a given class label. Since some annotations are missing in ImageNet-1k training images, we test these methods on the validation/test set of ImageNet-1k and CUB-200 to choose a better method to generate pseudo bounding boxes for PSOL. For the DDT method, we first resize the training images to the resolution size of $448\times448$, then perform DDT on training images. According to the statistics collected on training images, we generate bounding boxes on test images with the correct class label. For other WSOL methods, we follow original instructions in their papers and use pre-trained models to generate bounding boxes on validation/test images with the correct class label.

\begin{table*}
	\caption{Empirical localization accuracy results on CUB-200 and ImageNet-1k. The first column of the paper shows the model name, and the second column shows the backbone network for each model. Parameter number and FLOPs are shown in the third and fourth column. Then \textit{Top-1/Top-5 Loc} accuracy of CUB-200 and ImageNet-1k are shown in the next four columns. The last column illustrates the \textit{GT-Known Loc} accuracy on ImageNet-1k. For separate models like DDT and our \texttt{-Sep} models, we combine their localization results with classification results of baseline models. For FLOPs calculation, we only calculate convolutional operations as FLOPs and using networks on ImageNet as counting examples. Results with bold are best among the same backbone networks.}
	\label{table:all clsloc results}
	\setlength{\tabcolsep}{1.5pt}
	\centering
	\begin{tabular}{|l|r|r|r|r|r|r|r|r|}
		\hline
		\multirow{2}{*}{Model}  & \multirow{2}{*}{Backbone} & \multirow{2}{*}{Parameters} & \multirow{2}{*}{FLOPs} & \multicolumn{2}{|c|}{CUB-200}&\multicolumn{3}{|c|}{ImageNet-1k}\\ 
		\cline{5-6} \cline{7-9}
		& & &  & Top-1 Loc & Top-5 Loc & Top-1 Loc &Top-5 Loc & GT-Known Loc\\
		\hline 
				
		VGG16-CAM~\cite{camcvpr2016} & VGG-GAP & 14.82M & 15.35G & 36.13 & - & 42.80 & 54.86 & 59.00\\
		VGG16-ACoL~\cite{acolcvpr2018} & VGG-GAP & 45.08M & 43.32G  & 45.92 & 56.51 & 45.83 & 59.43 & 62.96\\
		ADL~\cite{adlcvpr2019} & VGG-GAP & 14.82M & 15.35G  & 52.36 & - & 44.92 & - & -\\
		VGG16-Grad-CAM~\cite{gradcamiccv2017} & VGG16 & 138.36M & 15.42G &- & -& 43.49 & 53.59 & -\\
		CutMix~\cite{cutmixiccv2019} & VGG-GAP & 138.36M & 15.35G & 52.53 & - & 43.45 & - &- \\
		DDT-VGG16~\cite{ddtpr2019} & VGG16 &  138.36M & 15.42G & 62.30 & 78.15 & 47.31 & 58.23 & 61.41\\
		\hline 
		PSOL-VGG16-Sep & VGG16 & 274.72M & 30.83G & \textbf{66.30} & \textbf{84.05} & \textbf{50.89} & \textbf{60.90} & \textbf{64.03}\\ 
		PSOL-VGG16-Joint & VGG16 & 140.46M & 15.42G & 60.07 & 75.35 & 48.83 & 59.00 & 62.1\\
		PSOL-VGG-GAP-Sep & VGG-GAP & 29.64M & 30.70G & 59.29 & 74.88 & 48.36 & 58.75 & 63.72\\ 
		PSOL-VGG-GAP-Joint & VGG-GAP & 15.08M & 15.35G & 58.39 & 72.64 & 47.37 & 58.41 & 62.25\\
		\hline \hline 
		SPG~\cite{spgeccv2018} & InceptionV3 & 38.45M & 66.59G & 46.64 & 57.72 & 48.60 & 60.00 & 64.69\\
		ADL~\cite{adlcvpr2019} & InceptionV3 & 38.45M & 66.59G & 53.04 & - & 48.71 & - & -\\
		\hline 
		PSOL-InceptionV3-Sep & InceptionV3 & 53.32M & 11.42G & \textbf{65.51} & \textbf{83.44} & \textbf{54.82} & \textbf{63.25} & \textbf{65.21} \\
		PSOL-InceptionV3-Joint & InceptionV3 & 29.21M & 5.71G & 60.32 & 78.98 & 52.76 & 61.10 & 62.83\\
		\hline \hline 
		ResNet50-CAM~\cite{camcvpr2016} & ResNet50 & 25.56M & 4.10G & 29.58 & 37.25 & 38.99 & 49.47 & 51.86 \\
		ADL~\cite{adlcvpr2019} & ResNet50-SE & 28.09M & 6.10G & 62.29 & - & 48.53 & - & -\\
		CutMix~\cite{cutmixiccv2019} & ResNet50 & 26.61M & 4.10G & 54.81 & - & 47.25 & - & - \\
		\hline 
		PSOL-ResNet50-Sep & ResNet50 & 50.12M & 8.18G & \textbf{70.68} & \textbf{86.64} & \textbf{53.98} & \textbf{63.08} & \textbf{65.44}\\
		PSOL-ResNet50-Joint & ResNet50 & 26.61M & 4.10G & 68.17 & 83.69 & 52.82 & 62.00 & 64.30\\
		\hline \hline
		DenseNet161-CAM & DenseNet161 & 29.81M & 7.80G & 29.81 & 39.85  & 39.61 & 50.40 & 52.54 \\
		PSOL-DenseNet161-Sep & DenseNet161 & 56.29M& 15.46G & \textbf{74.97} & \textbf{89.12} & \textbf{55.31} & \textbf{64.18} & \textbf{66.28} \\
		PSOL-DenseNet161-Joint & DenseNet161 & 29.81M & 7.80G & 74.24 & 87.03 & 54.48 & 63.41 & 65.39\\
		\hline
	\end{tabular}
\end{table*} 

We list the \textit{GT-Known Loc} of DDT and weakly supervised localization methods in Table~\ref{table:gtknown results}. As shown in Table~\ref{table:gtknown results}, DDT achieves comparable results with WSOL methods on ImageNet-1k, but achieves better performance than all WSOL methods on CUB-200. DDT results on CUB-200 indicate that object localization should not be related to classification labels. Furthermore, these WSOL methods need large computational resources, e.g., storing the feature map of each image, then perform off-line CAM operation to get the final bounding boxes. Compared to these methods, DDT has little computation requirements and achieves comparable results. For the base model choice of DDT, though DDT-DenseNet161 has higher accuracy than DDT-VGG16 on ImageNet-1k, it runs much slower due to the dense connections and has lower accuracy than DDT-VGG16 on CUB-200. Based on these observations, we choose DDT with VGG16 to generate our bounding boxes on training images in PSOL. 

\subsection{Comparison with State-of-the-art Methods}

In this section, we will compare our PSOL models with state-of-the-art WSOL methods: CAM~\cite{camcvpr2016}, HaS~\cite{hideandseekiccv2017}, ACoL~\cite{acolcvpr2018}, SPG~\cite{spgeccv2018} and ADL~\cite{adlcvpr2019} on CUB-200 and ImageNet-1k.

\begin{figure*}
	\centering
	\subcaptionbox{CUB-200-2011}{\includegraphics[width=1.0425\columnwidth]{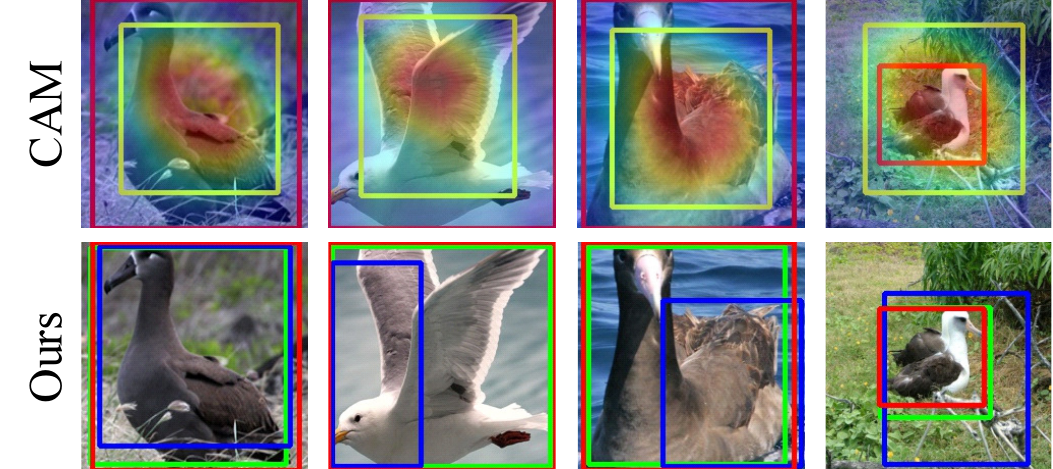}}
	\subcaptionbox{ImageNet-1k}{\includegraphics[width=1.0425\columnwidth]{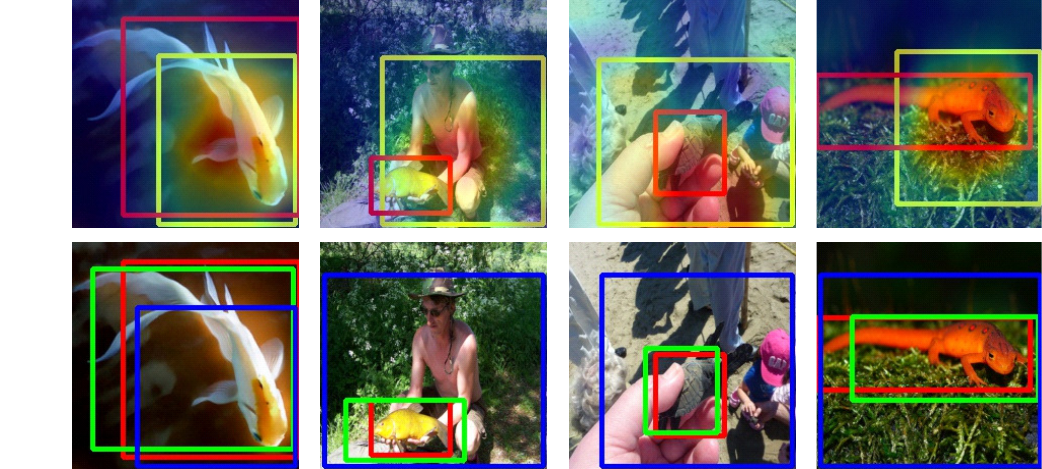}}
	\caption{Comparison of our methods with CAM and DDT. Please note that in CAM figures, yellow boxes are CAM predicted boxes and red boxes are groundtruth boxes. In figures of our methods, blue boxes are DDT generated boxes, green boxes are predicted boxes by our regression model and red boxes are groundtruth boxes. We use the DenseNet161-Sep model to output DDT and predict boxes. This figure is best viewed in color and zoomed in. }
	\label{fig:examples}
\end{figure*}

We list experimental results in Table~\ref{table:all clsloc results}. Furthermore, we visualize bounding boxes generated by CAM~\cite{camcvpr2016}, DDT~\cite{ddtpr2019} and our methods in Fig.~\ref{fig:examples}. According to these results, we have the following findings.

\begin{itemize}
	\item Without any training, DDT already performs well on both CUB-200 and ImageNet. DDT-VGG16 achieves 47.31\% \textit{Top-1 Loc} accuracy, which has a 2$\sim$3\% edge over WSOL models based on VGG16. Since DDT is a class-agnostic method, it suggests that WSOL should be divided into two independent sub-tasks: class-agnostic object localization and object classification.
	\item All PSOL models with separate training perform better than PSOL models with joint training. In all five baseline models, -Sep models consistently perform better than -Joint models by large margins. These results indicate that learning with joint classification and localization is not suitable.
	\item All our PSOL models enjoy a large edge~(mostly $>5$\%) on CUB-200 compared with state-of-the-art WSOL methods, including the DDT-VGG16 method. CUB-200 is a fine-grained dataset which contains many categories of birds. The within-class variation is much larger than the between-class variation in most fine-grained datasets~\cite{cubtech2011}. The exact label may not help the process of localization. Hence, the co-localization method DDT will perform better than previous WSOL methods.
	\item CNN has the ability to tolerate some incorrect annotations, and retain high accuracy on validation sets. For all separate localization models, the \textit{GT-Known Loc} is higher than DDT-VGG16. This phenomenon indicates that CNN can tolerate some annotation errors and learn robust patterns from noisy data. 
	\item Some restrictions and rules-of-thumb in WSOL do not carry over to PSOL. In previous WSOL papers, only one final fully connected layer is allowed, and large spatial size of the output feature map is recommended. Many methods try to remove the stride of the last downsample convolution layer, which will result in large FLOPs~(such as SPG and VGG16-ACoL). Besides this, three fully connected layers in VGG16 are all removed, which will directly affect the accuracy. However, in our experiments, VGG-Full performs significantly better than VGG-GAP. Since CAM requires GAP and only one FC layer, when this restriction is removed, VGG16 can get better performance. Another restriction is the inference path of the network. WSOL needs the output of the last convolutional layer in the model, and often uses simple forward networks~(VGG16, GoogLeNet, and InceptionV3). Complex network structures like DenseNet are not recommended and do not perform well in the WSOL problem~\cite{camcvpr2016}. As we show in Table~\ref{table:all clsloc results}, CAM achieves poor performance with DenseNet161. DenseNet will use features of every block, not just the last feature to conduct classification. Thus, the semantic meaning of the last feature may not as clear as the last feature of sequential networks like ResNet and VGG. However, PSOL-DenseNet models are directly trained on noisy bounding boxes, which can avoid this problem. Moreover, DenseNet161 achieves the best performance.
\end{itemize}

\subsection{Transfer Ability on Localization}
In this section, we will discuss the transferability of different localization models. 

Previous weakly supervised localization models need the exact label to generate bounding boxes, regardless of the correctness of the label. However, our proposed method does not need the label and directly generate the bounding box. So we are interested in that: Is single object localization task transferable? Does the model trained directly on object localization tasks like trained on image recognition tasks, have good generalization ability?

We perform the following experiment. We take object localization models trained on ImageNet-1k, then predict on CUB-200 test images directly, i.e., without any training or fine-tuning process. We add previous WSOL methods for a fair comparison. Since they need exact labels, we fine-tune all these models. For all models marked with *, they are only fine-tuned with classification parts~(the last fully connected layer), i.e.,  features learned on ImageNet-1k are directly transferred to CUB-200. For models marked without *, they are fine-tuned on CUB-200 with all layers. We take our VGG-GAP-Sep model for fair comparison and DenseNet161-Sep model for better results. Results are in Table~\ref{table:transfer results}. 

\begin{table}
	\small
	\caption{Transfer results of our models on CUB-200 and ImageNet-1k. For a fair comparison, we add VGG-GAP with CAM, VGG16-ACoL~\cite{acolcvpr2018} and SPG~\cite{spgeccv2018} for transfer experiments. VGG-GAP is fine-tuned with all layers while VGG-GAP* only fine-tunes the final fully connected layer. Please note that PSOL models trained on ImageNet-1k do not have any training or fine-tuning process on CUB-200.} 
	\label{table:transfer results}
	\setlength{\tabcolsep}{2pt}
	\centering
	\begin{tabular}{|l|r|r|r|}
		\hline
		Model   & Trained & Target & GT-Known Loc\\
		\hline 
		VGG-GAP + CAM & CUB-200 & CUB-200 & 57.96 \\
		VGG-GAP* + CAM & ImageNet & CUB-200 & 57.53 \\
		VGG16-ACoL + CAM & CUB-200 & CUB-200 & 59.30 \\
		VGG16-ACoL* + CAM& ImageNet & CUB-200 & 58.70 \\
		SPG + CAM& CUB-200 & CUB-200 & 60.50 \\
		SPG* + CAM & ImageNet & CUB-200 & 59.70 \\
		\hline 
		PSOL-VGG-GAP-Sep & CUB-200 & CUB-200 & 80.45\\
		PSOL-VGG-GAP-Sep & ImageNet & CUB-200 & 89.11\\
		\hline 
		PSOL-DenseNet161-Sep & CUB-200 & CUB-200 & 92.54 \\
		PSOL-DenseNet161-Sep & ImageNet & CUB-200 & 92.07 \\
		\hline
	\end{tabular}
\end{table}

It is surprising that without any supervision, PSOL object localization models can transfer well from ImageNet-1k to CUB-200, which performs significantly better than previous WSOL methods, including models which only fine-tune the classification weight~(models marked with *), and models which fine-tune the whole weights. It further indicates that objection localization is not dependent on classification, and it is inessential to perform object localization with the class label. Furthermore, it proves the advantage of our PSOL method.
\subsection{Combining with State-of-the-art Classification}
Previous methods try to combine localization outputs with state-of-the-art classification outputs to achieve better localization results. SPG~\cite{spgeccv2018} and ACoL~\cite{acolcvpr2018} combine with DPN networks including DPN-98, DPN-131 and DPN-ensemble~\cite{dpnnips2017}. For a fair comparison, we also combine other models'~(InceptionV3 and DenseNet161) results with DPN-131. Moreover, EfficientNet~\cite{efficientneticml2019} achieves better results on ImageNet-1k recently. We combine our localization outputs with EfficientNet-B7's classification outputs. Results are in Table~\ref{table:combine results}.

\begin{table}
	\small	
	\centering
	\caption{Top-1  and Top-5 Loc results by combining localization of our models with more state-of-the-art classification models on ImageNet-1k.}
	\label{table:combine results}
	\setlength{\tabcolsep}{2pt}
	\begin{tabular}{|l|r|r|}
		\hline
		Model   & Top-1 & Top-5 \\
		\hline 
		VGG16-ACoL+DPN131   & 53.94 & 61.15 \\
		VGG16-ACoL+DPN-ensemble  & 54.86 & 61.45 \\
		SPG + DPN131  & 55.19 & 62.76 \\
		SPG + DPN-ensemble  & 56.17 & 63.22 \\
		\hline 
		PSOL-InceptionV3-Sep + DPN131  & 55.72 & 63.64 \\				
		PSOL-DenseNet161-Sep + DPN131  & 56.59 & 64.63 \\
		PSOL-InceptionV3-Sep + EfficientNet-B7  & 57.25 & 64.04 \\				
		PSOL-DenseNet161-Sep + EfficientNet-B7  & 58.00 & 65.02 \\
		\hline
	\end{tabular}
\end{table}

From the table we can see that our model achieves better localization accuracy on ImageNet-1k compared with SPG~\cite{spgeccv2018} and ACoL~\cite{acolcvpr2018} when combining the same classification results from DPN131~\cite{dpnnips2017}. Furthermore, when combining with EfficientNet-B7~\cite{efficientneticml2019}, we can achieve 58.00\% Top-1 localization accuracy.
\subsection{Comparison with fully supervised methods}
We also compare our PSOL with fully supervised localization methods on ImageNet-1k. Fully supervised methods use training images with accurate bounding box annotations in ImageNet-1k to train their models. Results are in Table~\ref{table:fully supervised results}.

\begin{table}
	\small
	\centering
	\caption{Compare our method with state-of-the-art fully supervised methods on ImageNet-1k validation datasets.}
	\label{table:fully supervised results}
	\setlength{\tabcolsep}{2pt}
	\begin{tabular}{|l|r|r|}
		\hline
		Model   &  supervision &Top-5 Loc\\
		\hline 
		GoogLeNet-GAP~\cite{camcvpr2016} & weak & 57.1 \\
		GoogLeNet-GAP~(heuristics)~\cite{camcvpr2016} & weak & 62.9 \\
		VGG16-Sep & weak & 60.9 \\
		DenseNet161-Sep & weak &  64.2 \\
		\hline 
		GoogLeNet~\cite{googlenetcvpr2015} & full & 73.3 \\
		OverFeat~\cite{overfeaticlr2014} & full & 70.1 \\
		AlexNet~\cite{alexnetnips2012} & full & 65.8 \\
		VGG16~\cite{vggiclr2014} & full & 70.5 \\
		VGGNet-ensemble~\cite{vggiclr2014} & full & 73.1\\
		ResNet + Faster-RCNN-ensemble~\cite{fasterrcnnnips2015} & full & 90.0 \\
		\hline
	\end{tabular}
\end{table}

With the bounding box regression sub-network, our DenseNet161-Sep model can roughly match fully supervised AlexNet with Top-5 Loc accuracy. However, our performances are still worse than fully supervised OverFeat, GoogLeNet and VGGNet. It is noticeable that ResNet + Faster-RCNN-ensemble~\cite{fasterrcnnnips2015} achieves the best Top-5 Loc accuracy. They transfer region proposal networks trained on ILSVRC detection track, which has 200 classes of fully labeled images, to the 1000-class localization tasks directly. The region proposal network shows good generalization ability among different classes without fine-tuning, which indicates that localization is separated with classification.
\section{Discussions and Conclusions}
In this paper, we proposed the pseudo supervised object localization~(PSOL) to solve the drawbacks in previous weakly supervised object localization methods. Various experiments show that our methods obtain a significant edge over previous methods. Furthermore, our PSOL methods have good transfer ability across different datasets without any training or fine-tuning.

For future works, we will try to dive deep into the joint classification and localization problem: We will try to integrate both tasks into a single CNN model with less localization accuracy drop. Another direction is trying to improve the quality of generating bounding boxes with class-agnostic methods. Finally, novel network structures or algorithms on localization problems should be found, which should prevent the high input resolution and computational resources in the current detection framework to apply to large-scale datasets.

{\small
\bibliographystyle{ieee_fullname}
\bibliography{egbib}
}

\end{document}